# FTA-FTL: A Fine-Tuned Aggregation Federated Transfer Learning Scheme for Lithology Microscopic Image Classification

Keyvan RahimiZadeh [a, *], Ahmad Taheri [b], Jan Baumbach [b], Esmael Makarian [c], Abbas Dehghani [d], Bahman Ravaei [d], Bahman Javadi [e], Amin Beheshti [a]

[a] School of Computing, Macquarie University, Sydney, NSW, 2109, Australia
[b] Institute of Computational Systems Biology, University of Hamburg, Notkestrasse 9, 22607 Hamburg, Germany
[c] Department of Mining Engineering, Sahand University of Technology, Tabriz 513351996, Iran
[d] Department of Computer Engineering, School of Engineering, Yasouj University, Yasouj 7591874934, Iran
[e] School of Computer, Data and Mathematical Sciences, Western Sydney University, NSW 2751, Australia

**Abstract:** Lithology discrimination is a crucial activity in characterizing oil reservoirs, and processing lithology microscopic images is an essential technique for investigating fossils and minerals and geological assessment of shale oil exploration. In this way, Deep Learning (DL) technique is a powerful approach for building robust classifier models. However, there is still a considerable challenge to collect and produce a large dataset. Transfer-learning and data augmentation techniques have emerged as popular approaches to tackle this problem. Furthermore, due to different reasons, especially data privacy, individuals, organizations, and industry companies often are not willing to share their sensitive data and information. Federated Learning (FL) has emerged to train a highly accurate central model across multiple decentralized edge servers without transferring sensitive data, preserving sensitive data, and enhancing security. This study involves two phases; the first phase is to conduct Lithology microscopic image classification on a small dataset using transfer learning. In doing so, various pre-trained DL model architectures are comprehensively compared for the classification task. In the second phase, we formulated the classification task to a Federated Transfer Learning (FTL) scheme and proposed a Fine-Tuned Aggregation strategy for Federated Learning (FTA-FTL). In order to perform a comprehensive experimental study, several metrics such as accuracy, f1 score, precision, specificity, sensitivity (recall), and confusion matrix are taken into account. The results are in excellent agreement and confirm the efficiency of the proposed scheme, and show that the proposed FTA-FTL algorithm is capable enough to achieve approximately the same results obtained by the centralized implementation for Lithology microscopic images classification task.

## 1- Introduction

Reservoir characterization is one of the most crucial tasks in oil and gas studies. In addition, it is an important aspect of determining lithology because it significantly contributes to provide well planning from exploration to enhanced oil recovery (EOR) stag (He et al., 2022). For example, porosity and permeability are critical parameters in a hydrocarbon system and deeply depend on lithology. Therefore, knowing the facies can be advantageous to forecasting the volume of hydrocarbon in place and lessening uncertainty over reservoir characterization (Liu et al., 2022). Additionally, lithology is an integral part of geotechnical and rock physics models and is crucial for drilling and production operations (Johari and Emami Niri, 2021). A lack of correct diagnosis can face petroleum engineers and geologists serious problems such as sand production, collapse, and eruption because each lithology presents specific mechanical and geochemical features that must be recognized separately (Cai et al., 2021). In the EOR and carbon capture and storage (sequestration) (CCS) processes, lithology is one of the primary forces and the parameter that is investigated, firstly. The former is performed to increase production in oil and gas reservoirs in the last half of their life, and the latter is a new technology for storing carbon dioxide ($Co_2$) in the underground reservoirs to reduce carbon footprint in the environment (Tan et al., 2022). A part from petroleum geosciences, in other fields related to earth science such as mining exploration, geodesy, and civil engineering, lithology is a focal factor that must be identified quickly and accurately (Zhang et al., 2017).

---

* Corresponding author
E-mail address: keyvan.rahimizadeh@mq.edu.au (K. RahimiZadeh), ahmad.taheri@uni-hamburg.de (A. Taheri), jan.baumbach@uni-hamburg.de (J. Baumbach), esmael.makarian@gmail.com (E. Malekian), dehghani@yu.ac.ir (A. Dehghani), ravaei@yu.ac.ir (b. Ravaei), b.javadi@westernsydney.edu.au (B. Javadi), amin.beheshti@mq.edu.au (A. Beheshti).

There are various methods for Lithology discrimination, including direct and indirect approaches. Laboratory-based core analysis is the most common and reliable method for directly identifying different lithologies. In these methods, by using a rotary core catcher, core samples are taken from the side of drilled wells, maintained intact, and transferred to the surface for manual examination, such as thin–sections. A thin layer of taken cores under geological optical microscopes, minerals, and fossils are studied in these sections (Xu et al., 2022).

Moreover, the lithology of different formations in laboratories is recognized by modern technology facilities such as scanning by electron microscope (SEM), CT scan, and Electron Probe Micro-Analyzer (EPMA) with a focus on minerals of rock and their constituent elements (Izadi et al., 2017). These methods demand a great deal of time and money, so thin sections are more used in laboratories as a cost-effective and reliable method for lithology identification (Xu et al., 2022). Extensive research has long been performed by alternative methods such as geophysical log analysis (well-logging), seismic data, rock physics, numerical modeling, and especially geostatistical methods (Al-Mudhafar, 2017; Johari and Emami Niri, 2021; McCreery and Al-Mudhafar, 2017; Wang et al., 2022). Performing myriad projects by petroleum geologists and engineers, these methods have some issues. For instance, establishing a relationship between lithology and seismic data would be difficult (Ren et al., 2022). Moreover, well–logging operation is challenging or maybe impossible in all situations (e.g., horizontal wells), and well–logging devices may be error-prone in some situations (e.g., well casing) with undesirable effects on the results, (Liu et al., 2022).

With extensive improvements in Machine Learning (ML), researchers have shown a great inclination to employ various ML techniques for they are fast, cost-effective, and versatile. Above all, these methods are capable of dealing with big data analysis (Hussain et al., 2022). A brief review of the recent related studies on lithology identification is presented in Table 1. Since these studies did not pay special attention to predicting lithology using images from cores, it is needed to conduct research on lithology types determination based on images captured from cores. Nowadays, many ML techniques are utilized to create robust image classifier models. Among all of them, DL methods have shown a significant performance; however, there are several challenges to training a robust DL-based classifier model, such as computational resource limitations, time constraints, and especially the most crucial limitation is the lack of enough data samples (Olivas et al., 2009). It is useful to employ transfer learning as an efficient and practical approach to surmount the challenges as mentioned earlier. Transfer learning refers to the learning enhancement in a new task by using the knowledge transferred from an already learned related task (Bozinovski, 2020). In addition, due to different reasons, especially data privacy, individuals, organizations, and industry companies often are not willing to share their sensitive data and information. Therefore, it is crucial to employ techniques to preserve sensitive data and enhance security issues. Federated Learning (FL) has emerged as a paradigm to train a global model with decentralized datasets in a federated manner. FL can train a highly accurate central model across multiple decentralized edge servers without transferring sensitive data.

In this study, firstly, we apply a central deep learning model architecture using transfer learning to classify/identify different lithologies from thin sections and microscopic studies in a carbonate petroleum reservoir. We perform comprehensive experiments on the state-of-the-art DL model architectures such as DensNet169 (Huang et al., 2017), Xception (Chollet, 2017), ResNet101 (He et al., 2015), VGG19 (Simonyan and Zisserman, 2014), and Inception-ResNet v2 (Szegedy et al., 2016) as the base models and XGBoost Tree (Chen and Guestrin, 2016) and FC model as the classifiers to determine the suitable DL model architecture for the lithologies image classification task. Furthermore, the lithologies microscopic images dataset is developed in the current study. In this way, we apply data augmentation techniques to increase the number of samples and overcome the imbalance challenge of the dataset. Our proposed dataset would be publicly available for further research. In the second phase, to provide a secure and privacy-preserving approach for sensitive data and overcome the difficulty of collecting enough data samples, a Federated Transfer Learning (FTL) scheme is introduced to perform the Lithology microscopic image classification task. The experimental results show that the proposed FTL scheme is capable enough to train a global model with both Independent & Identically Distributed (IID) and non-IID data in a federated learning manner. The results are presented in terms of accuracy, f1 score, precision, specificity, sensitivity (recall), and confusion matrix. The rest of the paper is organized as follows: In section 2, the process of dataset generation and pre-processing will be explained. The preliminaries and methods are described in section 3. Furthermore, the experiment analysis and discussion will be presented in section 4. Finally, the conclusion is presented in section 5.

Table 1. A brief review on the recent related studies on lithology identification

| Authors | Methods | Dataset type |
|---|---|---|
| Glover *et al.*, 2022 | - 8 Machine Learning clustering approaches such as K-means and Expectation Maximization (EM) | A large petrophysical database on tight carbonate cores |
| He *et al.*, 2022 | - K-Nearest Neighbors (KNN), Decision Tree (DT), Random Forest (RF), and XGBoost <br> - Bootstrap sampling is utilized | Well – logging |
| Xu et al., 2022 | - Xception, MobileNet_v2, Inception_ResNet_v2, Inception_v3, Densenet121, ResNet101_v2, and ResNet-101 | Microscopic images |
| Fu *et al.*, 2022 | - ResNeSt-50 <br> - channel-wise attention, multi-path network and transfer learning | Drill Macroscopic core images |
| Galdames *et al.*, 2022 | - CNN <br> - Transfer learning | Hyperspectral images |
| Liu *et al.*, 2022 | - One-dimensional Residual Network <br> - different regularization | Seismic data |
| Taheri *et al.*, 2022 | - AGMDH-type neural networks and Metaheuristic Algorithms | Well -logging |
| Ren *et al.*, 2022 | - Fuzzy decision tree and k-means | Well – logging |
| Xu *et al.*, 2021 | - ResNet-50 architecture <br> - Data augmentation to improve generalization | Macroscopic Images |
| Anjos *et al.*, 2021 | - CNN with a spatial pyramid and a global average pooling layer | Micro-CT images |
| Xu *et al.*, 2021 | - A multi-path network, Convolutional and Fully-connected | Microscopic images and rock element data |
| Chen *et al.*, 2020 | - CNN <br> - Mobilenet and ResNet architectures | Drilling String Vibration Data |
| Liu *et al.*, 2020 | - A transfer learning mechanism called the data drift joint adaptation extreme learning machine (DDJA-ELM) | Well -logging |
| Imamverdiyev *et al.*, 2019 | - A new 1D CNN based model | Well -logging |

**2- Dataset Generation and Pre-Processing**

In this study, a mini dataset of lithology microscopic images is generated for both multi-class and binary classification task. In the following, the process of producing the mini dataset is described in detail and step by step.

**2-1- Lithology Microscopic Images description**

Core studies provide an array of useful information on petrophysical features of the study area. The hydrocarbon formation studied comprises mainly oil and water without gas. The study of thin sections shows that there are three main lithologies, including Argillaceous Limestone, Limestone, and Dolomite, as shown in Fig. 1. Argillaceous Limestone is a type of sedimentary rock that includes a significant amount (but less than 50%) of clay comprising kaolinite, montmorillonite, illite, and chlorite. Limestone is another sedimentary rock that is largely comprised of calcium carbonate ($CaCO_3$), typically in the form of calcite or aragonite. Dolomite which widely recognized as Dolostone, is a sedimentary carbonate rock rich in $CaMg(CO_3)_2$. The lithology column effectively shows the intervals of rocks and how they are placed versus depth, which gives a better understanding of the studied area. The core samples are utilized in this study obtained from an oil carbonate reservoir in the Southern Persian Gulf. Regarding Fig. 1, Argillaceous Limestone is responsible for the most significant share of lithology; however, the contributions of Dolomite and Limestone are almost the same.

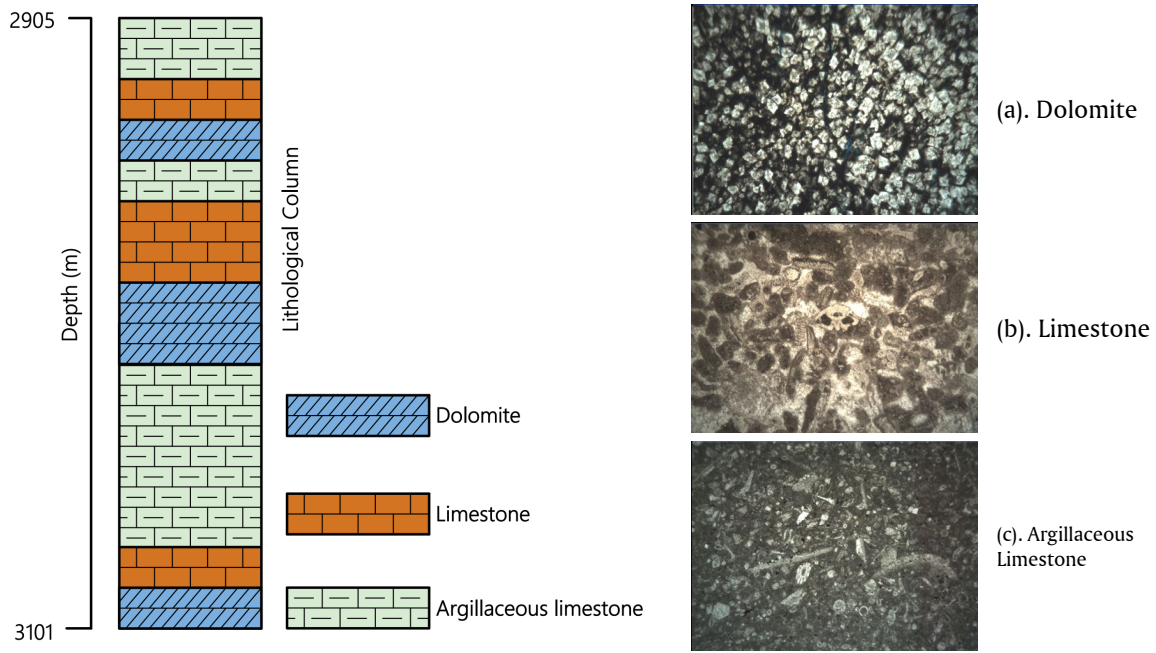

Figure 1. A schematic view of Lithology column of the understudy well and Lithofacies microscopic images

**2-2- Data augmentation**

Generally, the performance of most ML models, DL in particular, depends on the quality, quantity, and relevancy of training data samples to tackle the overfitting challenge and improve the model generalization (Goodfellow et al., 2017). However, labeled data for real-world applications may be limited in practice because collecting data might be time-consuming and costly in many project cases. In these cases, data augmentation techniques have become an important part of successful DL application on image data to enhance the diversity and sufficiency of training data set efficiently (Goodfellow et al., 2017; Yang et al., 2022). Fig. 2 and Table 2, show statistical information and data distribution of our dataset.

Data augmentation enhances the robustness of ML models by employing a wide range of techniques to increase the amount of data by artificially generating new data samples that the model might see in the real world based on the existing samples. It includes making small changes to data or using DL models to generate new data points.

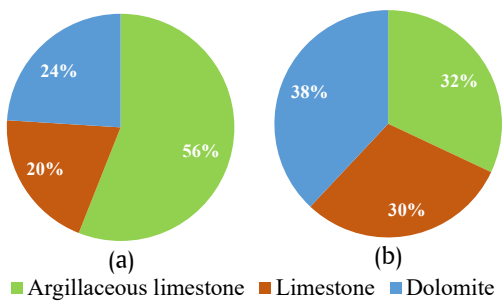

Figure 2. Percentage of data distribution in our dataset (a) original data before augmentation, (b) data after augmentation

Table 2. The number of images in each class before and after augmentation and class balancing process

| Classes | # Original | # Aug & Balanced |
|---|---|---|
| Argillaceous limestone | 684 | 684 |
| Limestone | 243 | 633 |
| Dolomite | 288 | 810 |

In our study, different data augmentation methods are applied to original image samples; specifically, methods such as flipping, scaling, cropping, segmentation, and rotation. In this way, we divide each original image into several segmentations (sub-images), as shown in Fig. 3, then the process of data augmentation is conducted to generate new samples. In addition, as our original image sample classes are imbalanced, we use augmentation techniques to solve these issues in our dataset by applying different augmentation rates for different classes.

### 2-3- Train and Test sets

As augmented datasets can significantly decrease the generalization of an ML model, it is essential to conduct controlled experiments to evaluate the performance of the model trained with such datasets (Goodfellow et al., 2017). Therefore, in order to split a dataset into train and test sets and perform controlled experiments, we select image samples for the test set, manually. In this way, approximately about 20 percent of image samples (440 samples) are selected as the test set, and 80 percent of image samples (1687 samples) are selected for the train set. The dataset* developed in this study will be made publicly available for further research.

### 3. Preliminaries and Method

This section describes the proposed FTL-based Lithology microscopic image classifier in detail. First, the concepts of transfer learning and federated learning will be explained. Then, the proposed FTL scheme for Lithology microscopic image classification task is described.

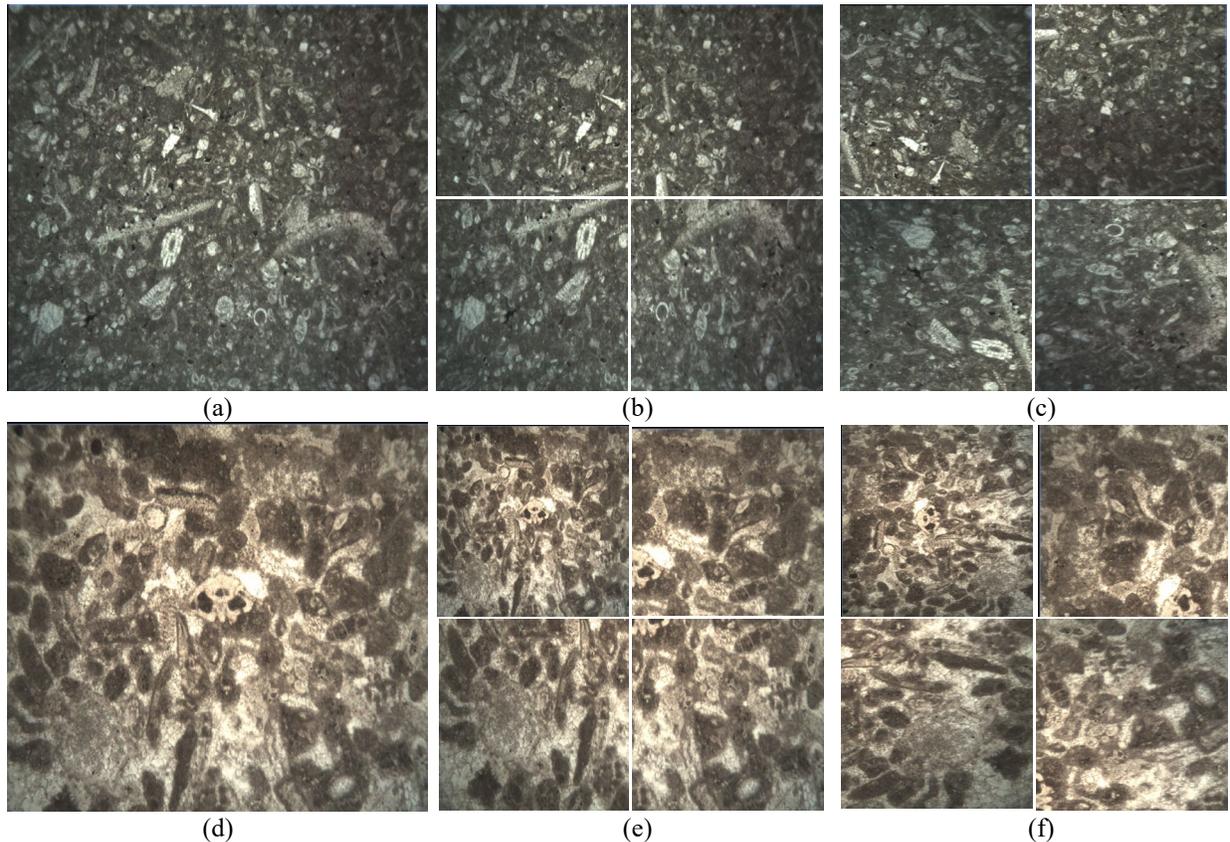

Figure 3. Examples of the data augmentation process used in this study: (a) and (d) original microscopic images, (b) and (e) segmented images, (c) rotated images by $90^0$, and (f) rotated images by $270^0$, after segmentation step.

---

* https://github.com/AhmadTaheri2021/Lithology-microscopic-images-mini-dataset.git

### 3.1. Transfer Learning

Human learners would seem to possess inherent ways to transfer knowledge between tasks. Facing new activities, we understand and use relevant knowledge based on prior learning experiences. The more related the new task is to our prior experiences, the more easily we can master it (Olivas et al., 2009). In contrast, isolated tasks are traditionally explained in the widely used ML algorithms. Transfer learning aims to change this by developing strategies for transferring knowledge gained in one or more source tasks and utilizing it to advance learning in a related target task (Zhang et al., 2017).

Due to the fact that the early layers' features/filters are more generic and the later layers are more original-dataset-specific in Convolutional Neural Network (CNN) models (Yamashita et al., 2018), it is reasonable to map the related task image classification to the early layers of a pre-trained CNN model and conduct transfer learning (Zhang et al., 2017). Subsequently, we use early layers of a pre-trained CNN-based image classifier to build our Lithology image classifier.

### 3-2- Federated Learning

Federated learning (FL) is a decentralized learning method that can allow several clients to train a global model without transferring local data off-site (McMahan et al., 2016). In FL, the majority of the training is done by the participants (clients), and a central server utilizes an aggregation approach to update the global model in an iterative process. The privacy-preserving nature of FL has been made it popular for a wide range of applications areas (Nasirigerdeh et al., 2020), for example, in Cross-block Oil-water Layer Identification (Chena et al., 2021), and healthcare data analysis (Chen et al., 2020; Sheller et al., 2018), where data access is restricted due to strict privacy policies.

Federated averaging (FedAvg) (Collins et al., 2022; McMahan et al., 2016) is a communication-efficient method for FL aiming to obtain an accurate global model with an efficient number of communication rounds between the server and clients. The main goal of FedAvg is to carry out a significant number of local updates in the clients before averaging the local model parameters on the server simply and weightedly. FedAvg can significantly decrease the communication rounds where the client's data has IID characteristics.

### 3-2- The Golden Section Search Algorithm

The Golden Section Search (GSS) is an effective method for locating extremum (minimum or maximum) using a sequential narrowing mechanism of the range of values inside that extremum exists (Agrawal and Aware, 2012). The GSS algorithm is more suitable for exploring the optimum of 1-Demintional unimodal functions (Taheri et al., 2021). The primary goal is to find the minimum value of function f(x) within the specific interval [$X_L$ $X_U$]. In doing so, in an iterative manner, two points $x_1$ and $x_2$ are chosen in the interval [$X_L$ $X_U$] and function f(x) is assessed at these points. Then, to shrink the search interval, as shown in Eq. (1), the point $X_L$ will be replaced by $X_1$ if f($x_1$) is better than f($x_2$), otherwise, the $X_U$ is replaced by $X_2$.

$$\begin{cases} x_L \leftarrow x_1 & \text{if } f(x_1) < f(x_2) \\ x_U \leftarrow x_2 & \text{otherwise.} \end{cases} \quad (1)$$

### 3-3- Model Architecture

The pre-trained Inception-ResNet v.2 is selected as the most suitable model for our purpose. Inception architecture has shown the capability to achieve excellent performance with a considerably low computational cost. Additionally, it has been proven that combining residual connections with the Inception architecture results in more performance improvement. It is because networks with residual Inception architecture outperform Inception networks without using residual connections. InceptionResNet v2 is a powerful hybrid residual Inception architecture network introduced in (Szegedy et al., 2016). We also utilize a dense layers (fully connected) architecture to design the classifier in this study.

Table 3. Symbols and description.

| Symbol | Description | Symbol | Description |
|---|---|---|---|
| $W^G$ | The global model | x | The decision variable |
| $W^H$ | The classifier | $X_L$ and $X_U$ | Upper and Lower bounds of decision variable x |
| $W^{Base}$ | The base model | $\varepsilon$ | Tolerance |
| $W^L$ | The local model (update) | $\varphi$ | Golden Ratio |
| $W^{Full}$ | The full model [$W^{Base} + W^H$] | $\sigma$ | Fine Ratio |
| $D^V$ | The validation data | k | The number of participants at each round. |
| T | The total number of rounds (iterations) | $n_k$ | The sample size of $k^{th}$ participant. |
| t | Each round (time) of global training | $N_t^{Total}$ | The total sample size of round t. |
| FC | Fully Connected | $\beta$ | A batch of data |
| $\eta$ | The local learning rate | E | The number local epochs |
| p | A sub-set of clients (participants). | | |

The Residual Inception Block is the fundamental building block of the architecture of Inception-ResNet v2. After each Inception block, a convolutional layer with 1×1 filter size is utilized to scale up the bank dimensionality of the expansion filter to match the input depth. Furthermore, convolutional filters with multiple-size and residual connections are incorporated in the Residual Inception Blocks. In addition, the batch normalization technique is only employed on top of the traditional layers of the Inception-ResNetV2. This training technique is used for significantly stabilizing the learning process and decreasing the number of training epochs that are required for training the Deep Neural Network DNN (Szegedy et al., 2016) .

Fig. 4. illustrates the proposed fine-tuned model of Inception-ResNet-V2 for Lithology microscopic image classification. The input layer in the CNN contains lithology microscopic images with the size of 299×299, and the network involves 164 layers. The Flatten layer converts the results of 2-Dimensional Avg-pooling arrays into a single long continuous linear vector.

3-4- Fine-Tuned Aggregation Federated Transfer Learning (FTA-FTL) Scheme

In this section, the proposed scheme of Fine-Tuned Aggregation Federated Transfer Learning, namely FTA-FTL, is presented. First, the process of the proposed FTL scheme is described step by step in Alg. 1. Then, our proposed FTA mechanism will be explained in detail, Alg. 2.

A schematic view of the architecture of the proposed FTL classification model is illustrated in Fig. 5. The process of the FTL scheme includes the server side and client side procedures. In this scheme, as shown in fig.5, the server side procedure acts as a central coordinator and hosted by cloud based infrastructure. According to Alg. 1, in this paradigm, decentralized learning is started by global model weights initialization (liens 1-2). As transfer learning is utilized, only the classifier is needed to be initialized. Then, in an iterative process (lines 3-15), the server communicates with clients to train the global model with distributed image datasets. In this way, a subset of clients is randomly selected to participate in the next round, in line 4. From lines 6 to 10, each participant downloads the full model $W_t^{G,Full}$ in a parallel fashion if it is the first round. Otherwise, only the classifier $W_t^{G,H}$ is required to be downloaded. After that, the client-side procedure is launched on each client to update/train the local model. In doing so, the global model weights are utilized as initial weights for the local model (line c.1). Before uploading the local update to the server, the process of training/optimization is done by using a stochastic gradient-based algorithm (lines c.2 – c.6). Then, in line c.7, the classifier is disjointed to be sent to the server. On the server, after collecting all local updates from participated clients, an aggregation process is conducted to produce the next generation of the global model using the GSS-based FTA Algorithm Alg. 2. (line 13). Finally, in line 14, the base model and global classifier are connected to generate a full model.

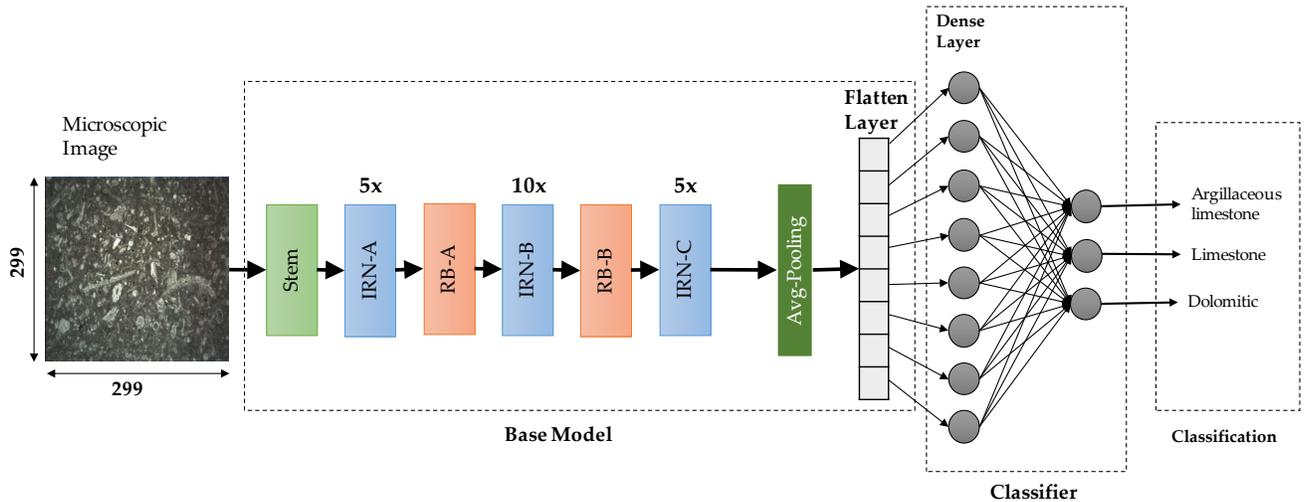

Figure 4. A schematic view of the proposed classification model architecture. IRN-A, IRN-B, and IRN-C are the Inception-ResNet block A, B and C, respectively. RB-A and RB-B are Reduction block type A and B, respectively.

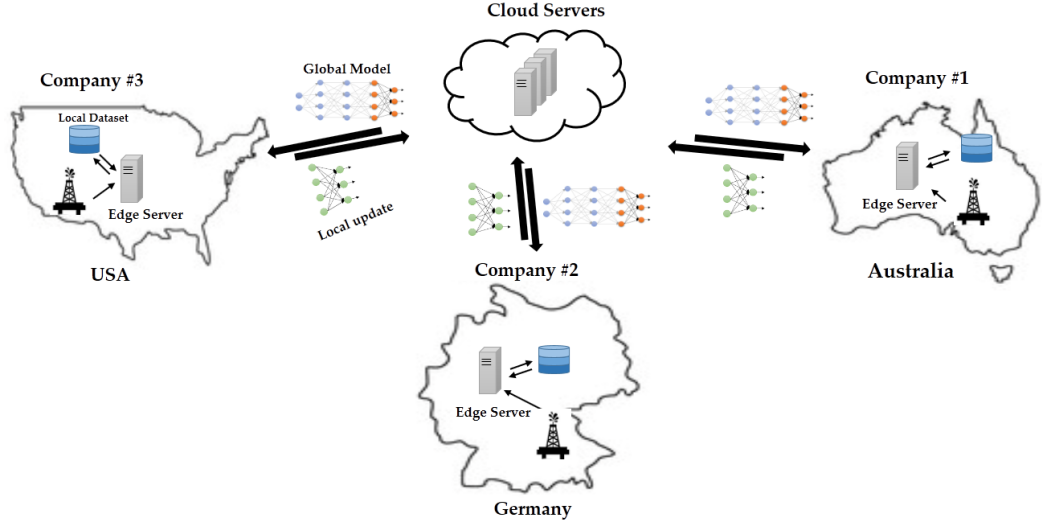

Figure 5. A schematic view of the architecture of the proposed Federated Transfer learning classification model.

In the GSS-based FTA Algorithm (Alg. 2), we employ a One-Dimensional Golden Section Search GSS algorithm (Agrawal and Aware, 2012) to optimize the aggregation process on the server. In doing so, after collecting local updates from participants at each round, the FTA algorithm is launched. According to Alg. 2, the FTA receives the global weights, local updates, and validation set as inputs. Then, in line 1, parameters lower bound $X_L$, upper bound $X_U$, and tolerance $\varepsilon$ are initialized. In line 2, the golden ration $\varphi$ is calculated. Then, in an iterative manner (lines 3-13), the GSS algorithm conducts a search process to find the minimum value of the objective function Eq. 2, with respect to the loss value of the aggregated global model on validation set in the interval [$X_L$,$X_U$]. The aim is to shrink the search interval. The search process is terminated when the interval [$X_L$,$X_U$] is lower than $\varepsilon$. After that, in lines 14 and 15, the fine ratio $\sigma$ and the next generation of global weights $W_{t+1}^{G,H}$ are calculated.

$$\min_{x \in \mathbb{R}^N, D^V \in D} F_X = L(W_{t+1}^G, D^V)$$
where:
$$W_{t+1}^G = W_t^G + \sigma \times \sum_{k=1}^{K} (\frac{n_k}{N_t^{Total}} \times (W_k^L - W_t^G)) \qquad (2)$$

where, $D^V$ is the validation set, $\sigma$ is the fine ratio estimated by GSS algorithm, and $L(W_{t+1}^G, D^V)$ is the loss function which calculates the loss value of the global model $W_{t+1}^G$ with $D^V$ and fine ratio.

---

**Algorithm 1. Federated Transfer Learning (FTL) algorithm**

*Server Side Procedure:*
1. Download Pre-trained base model $W^{Base}$ and freeze its layers
2. Initialize global classifier weights $W_0^{G,H}$
3. **for** each round $t = 1, 2, ..., T$ **do**
4.    Select a set of participants $P$ randomly // k number of clients
5.    **for** each Client $P_k$, k= 1, 2, 3, ...,K **do in parallel**
6.       **if** t == 1 or Client $P_k$ is a new participant **then**
7.          Client $P_k$ downloads full model $W_t^{G,Full}$, including $W^{G,Base}$ and $W_t^{G,H}$.
8.       **else**
9.          Client $P_k$ downloads the classifier $W_t^{G,H}$.
10.       **end if**
11.       $W_k^{L,H}, n_k \leftarrow \text{Update}_k(W_t^G)$
12.    **end for**
13.    **Aggregation**:
$$N_t^{Total} \leftarrow \sum_{k=1}^{K} n_k$$
       Get Global Weights $W_{t+1}^{G,H} \leftarrow \text{FTA}(W_t^G, W_t^{L,H}, D^V, N_t^{Total})$ // Alg. 2.
14.    Generate Full Model $W_t^{G,Full}$ by concatenating $W^{G,Base}$ and $W_t^{G,H}$
15. **end for**
16. **Return** $W^{G,Full}$
*End of Server Side Procedure*

*Client Side Procedure:*
1. Replace local model $W_k^L \leftarrow W_t^G$
2. **for** local epoch e from 1 to E do
3.    **for** batch $\beta \in \{\beta_1, \beta_2, \beta_3, ..., \beta_s\}$ **do**
4.       $W_{k,e+1}^L \leftarrow W_{k,e}^L - \eta \nabla L(W_{k,e}^L, \beta)$
         $n \leftarrow n + \|\beta\|$
5.    **end for**
6. **end for**
7. Split Classifier $W_k^{L,H}$
8. **Return** $W_k^{L,H}, n_k$
*End of Client Side Procedure*

**Algorithm 2. FTA: GSS based Fine-Tuned Aggregation (FTA) Algorithm**

**Inputs**: Global Weights, Local Updates, and Validation Set
**Outputs**: Aggregated Weights
1. **Initialize** $X_L$, $X_U$, and $\varepsilon$
2. **set:**

   Golden Ratio: $\varphi \leftarrow \dfrac{\sqrt{5}-1}{2}$

3. **While** $(x_U - x_L) > \varepsilon$ **do**
4.     $x_1 \leftarrow x_U - \dfrac{(x_U - x_L)}{\varphi}$
5.     $x_2 \leftarrow x_L + \dfrac{(x_U - x_L)}{\varphi}$
6.     $f_{x_1} \leftarrow F(x_1)$, Eq. (2)
7.     $f_{x_2} \leftarrow F(x_2)$, Eq. (2)
8.     **if** $f_{x_1} < f_{x_2}$ **then:**
9.         $x_U \leftarrow x_2$
10.     **else**
11.         $x_L \leftarrow x_1$
12.     **end if**
13. **end while**
14. **Calculate Fine Ratio** $\sigma \leftarrow \dfrac{(x_U + x_L)}{2}$
15. $W_{t+1}^{G,H} \leftarrow W_t^{G,H} + \sigma \times \sum_{k=1}^{K} \dfrac{n_k}{N_t^{Total}} \times (W_t^{L,H} - W_t^{G,H})$
16. **Return** $W_{t+1}^{G,H}$

## 4- Experimental Analysis

In this section, first, several experiments are conducted to investigate the performance of the state-of-the-art DNN model architectures such as DensNet169, Xception, ResNet10, VGG19, and Inception-ResNet v2, a brief review on the DL model architectures is presented in Table 4, and the results are presented in terms of accuracy, f1 score, precision, specificity, sensitivity (recall), and confusion matrix. Then, the best DNN classifier model is selected according to the results and utilized as the global model in our proposed FTL scheme.

Table 4. A brief review on the DL model architectures

| Model | Authors | Number of parameters (Million) | Number of layers (Network Depth) |
|---|---|---|---|
| ResNet101 | He et al., 2015 | 42.5 | 101 |
| Xception | Chollet, 2017 | 20.8 | 71 |
| Inception-ResNet V2 | Szegedy et al., 2016 | 54.3 | 164 |
| DenseNet169 | Huang et al., 2017 | 12.5 | 169 |
| VGG19 | Simonyan and Zisserman, 2014 | 139.6 | 19. |

## 4-1- Performance Metrics

The following metrics are utilized to analyze the experimental results. Accuracy is the metric that represents the portion of the correct predictions to total predictions, Eq. (3). When both False Negatives (FN) and False Positives (FP) are important, and the dataset is balanced, the Precision metric determines the proportion of positives that are true positives, Eq. (4). Sensitivity (Recall) measures the True Positive (TP) to total actual positives ratio of data Eq. (5). This metric is important when identifying the positives is very important and is useful to be more confident of the predicted positives. Similarly, the negative accuracy is the specificity. It represents True Negative (TN) to the total actual negatives ratio of data, Eq. (6). This metric is used to cover all True Negatives TN. By F1-Score, both precision and recall are considered in this metric. F1-Score is an overall data science measure of a model's accuracy that takes into account precision and recall, Eq. (7). The higher the value of F1-Score, the higher the balance between Recall and Precision metrics. The confusion matrix is a common measurement for classification task evaluation. It can be utilized for both binary and multiclass classification tasks. In Fig. 6, an example of a binary classification confusion matrix is illustrated. As shown in Fig. 6, the confusion matrix summarizes TP, FP, TN, and False FN values.

|  |  | Predicted Values | |
|---|---|---|---|
|  |  | Positive | Negative |
| Actual Values | Positive | TP | FN |
|  | Negative | FP | TN |

Figure 6. Confusion matrix

$$\text{Accuracy} = \frac{TP + TN}{TP + TN + FP + FN} \quad (3)$$

$$\text{Precision} = \frac{TP}{TP + FP} \quad (4)$$

$$\text{Sensitivity(Recall)} = \frac{TP}{TP + FN} \quad (5)$$

$$\text{Specificity} = \frac{TN}{TN + FP} \quad (6)$$

$$\text{F1-Score} = \frac{2 \times (\text{Recall} \times \text{Precision})}{(\text{Recall} + \text{Precision})} \quad (7)$$

## 4-2- Implementation settings

For implementing all algorithms, the following settings are used. Python 3.7 is used as the implementation programing language, and Tensorflow 2.8 is employed as the ML library and framework. In addition, we utilize the Adam optimization algorithm with a learning rate of 0.001 and 500 epochs to train the dense layer-based classifiers (fully connected layers). We also use the following setting for the XGBoost classifier: maximum depth=3, learning rate=0.3, and the number of estimators=100. To conduct a fair comparison, all models are trained with the same train set consisting of 80% samples, and then they are applied to classify the same images test set (20% of samples).

## 4-3- Results and Discussion

To select the most suitable architecture for lithology microscopic image classification, first, we evaluate various state-of-the-art DNN architecture performances. The results are presented in terms of accuracy, f1 score, precision, specificity, and sensitivity. In adition, we employ the selected DNN architecture to use and setup our FTL scheme.

### 4-3-1- Multi-class Classification

According to Table 8, it is clear to be seen that the Inception-ResNet v2 as the base model and a fully connected (FC) classifier with 2 (200,100) Dense layers outperforms other competitors and achieves the best results for evaluation metrics such as accuracy 90.68%, f1 score 90.76%, precision 90.61%, specificity 96.15%. In addition, among all other models, DenseNet169+XGB architecture performs better and takes the second best in terms of accuracy of 88.4%, f1 score of 88.3%, precision of 88.4%, and specificity of 90.6%. However, for the metric of sensitivity obtains the best result by 97.5%.

### 4-3-2- Binary Classification

Furthermore, we apply Inception-ResNet v2 architecture to conduct binary classification tasks. In doing so, a binary classification task is done for Argillaceous Limestone, limestone, and Dolomite classes and the experimental results are illustrated in Tables 5, 6, and 7, respectively. According to the presented experimental results, the Inception-ResNet v2 as the base model and a fully connected classifier with two dense layers provides the higher performance in all tasks for all evaluation metrics. Inception-ResNet v2 + FC architecture obtains the best result on the Argillaceous Limestone binary classification with accuracy of 95.45%, f1 score of 94.41%, precision of 95.32%, specificity of 98.07%, and sensitivity of 89.14. In addition, to perform comprehensive comparisons, the results of binary classifications are presented by using confusion matrixes in Fig. 7.

Table 5. Binary classification results for Argillaceous Limestone

| Base Model | Classifier | Accuracy | F1-Score | Precision | Specificity | Sensitivity |
|---|---|---|---|---|---|---|
| IncepResNet v2 | XGB | 0.9136 | 0.89533 | 0.8971 | 0.9421 | 0.8449 |
| IncepResNet v2 | FC | 0.9545 | 0.9441 | 0.9532 | 0.9807 | 0.8914 |

Table 6. Binary classification results for Limestone

| Base Model | Classifier | Accuracy | F1-Score | Precision | Specificity | Sensitivity |
|---|---|---|---|---|---|---|
| IncepResNet v2 | XGB | 0.8431 | 0.8112 | 0.8290 | 0.92 | 0.6785 |
| IncepResNet v2 | FC | 0.9136 | 0.9004 | 0.9079 | 0.9366 | 0.8642 |

Table 7. Binary classification results for Dolomite

| Base Model | Classifier | Accuracy | F1-Score | Precision | Specificity | Sensitivity |
|---|---|---|---|---|---|---|
| IncepResNet v2 | XGB | 0.884 | 0.8768 | 0.8812 | 0.9219 | 0.8245 |
| IncepResNet v2 | FC | 0.9340 | 0.9301 | 0.9338 | 0.9591 | 0.8947 |

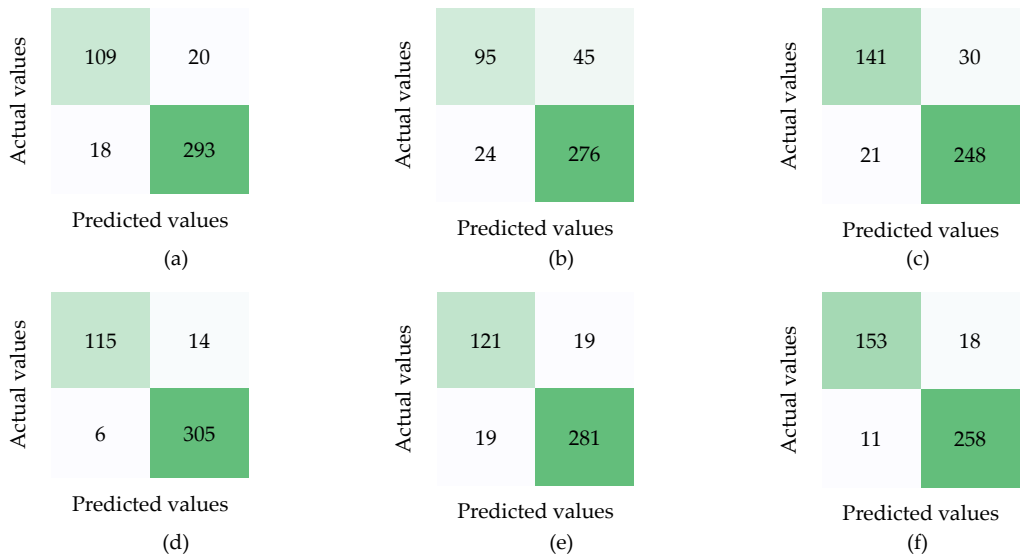

Figure 7. The confusion matrixes for binary classification. The first row (a-c) represents Inception-ResNet v2+XGB and the second row (d-f) represents Inception-ResNet v2+FC. (a) and (d): Argillaceous Limestone, (b) and (e): Limestone, (c) and (f): Dolomite.

### 4-3-3- FLT Scheme Results

To implement the FLT scheme, we use a straightforward and flexible federated deep-leaning simulator, namely, the FedSim[*] tool. In this way, the following settings are used in our experiments: the number of clients is set to 10, the maximum number of rounds is set to 100, and the optimization settings are the same as those used in the centralized manner. Furthermore, in this study, we utilize the FedAvg algorithm (McMahan et al., 2016) [35] as the base algorithm to develop our proposed FTL scheme for the lithology microscopic image classification task.

Table 9, presents the results of the proposed FTL scheme for multi-class Lithology image classification. It is clear from the results that the FTA-FTL algorithm is capable enough of achieving an acceptable result in multi-class Lithology image classification tasks with distributed data for both the IID and Non-IID cases. In addition, to conduct experimental analyses, the results of FLT classification are presented by using confusion matrixes, as shown in Fig. 8.

Table 8. Multi-class classification results

| Base Model | Classifier | Accuracy | F1-Score | Precision | Specificity | Sensitivity |
|---|---|---|---|---|---|---|
| Xception | XGB | 80.00 | 79.69 | 79.89 | 82.60 | 94.95 |
| Xception | FC | 82.50 | 82.34 | 82.43 | 87.50 | 93.27 |
| ResNet 101 | XGB | 64.77 | 64.09 | 64.65 | 74.46 | 83.03 |
| ResNet101 | FC | 65.95 | 65.60 | 65.72 | 76.21 | 84.43 |
| DenseNet169 | XGB | 88.40 | 88.31 | 88.39 | 90.62 | **97.52** |
| DenseNet169 | FC | 87.50 | 87.31 | 87.25 | 87.87 | 95.93 |
| VGG19 | XGB | 81.13 | 80.88 | 80.93 | 84.16 | 93.38 |
| VGG19 | FC | 81.13 | 81.21 | 81.32 | 89.14 | 91.20 |
| Inception-ResNet v2 | XGB | 86.81 | 86.50 | 86.81 | 91.05 | 90.98 |
| Inception-ResNet v2 | FC | **90.68** | **90.76** | **90.61** | **96.15** | 96.85 |

Table 9. Results of FTL scheme for multi-class classification

| Model | Data Distribution | Accuracy | F1-Score | Precision | Specificity | Sensitivity |
|---|---|---|---|---|---|---|
| FTL (FedAvg) | IID | 88.86 | 88.86 | 89.12 | 96.09 | 91.80 |
| FTL (FedAvg) | Non-IID | 87.72 | 87.77 | 87.76 | 91.91 | 96.82 |
| FTA-FTL | IID | 90.24 | 90.47 | 90.80 | 94.63 | 95.04 |
| FTA-FTL | Non-IID | 89.83 | 89.74 | 89.21 | 93.77 | 96.51 |
| DensNet169+XGB | Centralized | 88.40 | 88.31 | 88.39 | 90.62 | 97.52 |
| Inception-ResNet v2 +FC | Centralized | 90.68 | 90.76 | 90.61 | 96.15 | 96.85 |

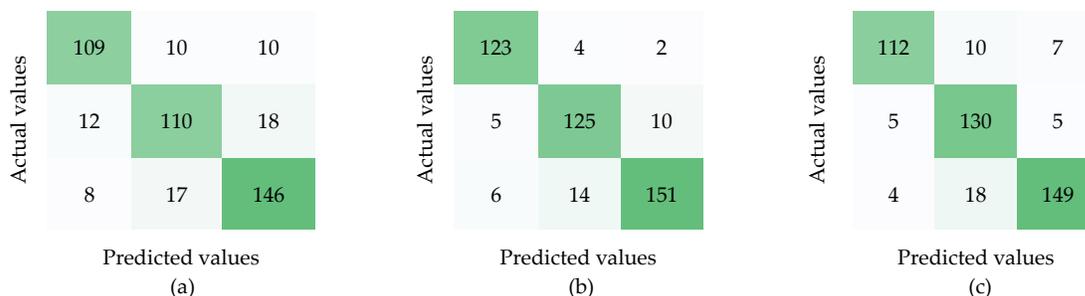

Figure 8. The Confusion matrixes for multi-class classification. (a). Inception-ResNet v2+XGB, (b). Inception-ResNet v2+FC, and (c) FTL.

---
[*] https://github.com/AhmadTaheri2021/Federated-Deep-Learning

## 5- Conclusion

In this study, a novel scheme was introduced for lithology microscopic image classification task. In this way, firstly, the performance of various DL model architectures was investigated in a central manner to determine the most suitable architecture for our purpose. After that, we proposed a Fine-Tuned Aggregation Federated Transfer Learning (FTA-FTL) scheme. In doing so, the classification task was formulated as a FTL scheme and a novel aggregation method named FTA was introduced and applied to enhance the convergence rate and communication costs of the proposed FTL. In addition, a Lithology microscopic images dataset was produced during this study. The results of this study confirmed that the proposed FTA-FTL scheme is able to train a global model with distributed data in the both IID and non-IID forms and obtains the higher accuracy than well-known federated learning algorithm (e.g. FedAvg) for test data. However, it is not expected to achieve a very high accuracy by FL based trained models, our proposed FTA-FTL scheme obtained approximately the same accuracy achieved by a centrally trained model.

## References

Agrawal, J., Aware, M., 2012. Golden section search (GSS) algorithm for Maximum Power Point Tracking in photovoltaic system, in: 2012 IEEE 5th India International Conference on Power Electronics (IICPE). IEEE, pp. 1–6. https://doi.org/10.1109/IICPE.2012.6450384

Al-Mudhafar, W.J., 2017. Integrating well log interpretations for lithofacies classification and permeability modeling through advanced machine learning algorithms. J. Pet. Explor. Prod. Technol. 7, 1023–1033. https://doi.org/10.1007/s13202-017-0360-0

Bozinovski, S., 2020. Reminder of the First Paper on Transfer Learning in Neural Networks, 1976. Informatica 44. https://doi.org/10.31449/inf.v44i3.2828

Cai, X., Guo, Q., Jiang, H., 2021. Utilizing Managed Pressure Drilling Technique to Prevent Rock Failure and Maintain Wellbore Stability. IOP Conf. Ser. Earth Environ. Sci. 861, 062055. https://doi.org/10.1088/1755-1315/861/6/062055

Chen, G., Chen, M., Hong, G., Lu, Y., Zhou, B., Gao, Y., 2020. A New Method of Lithology Classification Based on Convolutional Neural Network Algorithm by Utilizing Drilling String Vibration Data. Energies 13, 888. https://doi.org/10.3390/en13040888

Chen, T., Guestrin, C., 2016. XGBoost, in: Proceedings of the 22nd ACM SIGKDD International Conference on Knowledge Discovery and Data Mining. ACM, New York, NY, USA, pp. 785–794. https://doi.org/10.1145/2939672.2939785

Chen, Y., Qin, X., Wang, J., Yu, C., Gao, W., 2020. FedHealth: A Federated Transfer Learning Framework for Wearable Healthcare. IEEE Intell. Syst. 35, 83–93. https://doi.org/10.1109/MIS.2020.2988604

Chena, B., Zenga, X., Zhang, W., 2021. Federated Learning for Cross-block Oil-water Layer Identification.

Chollet, F., 2017. Xception: Deep Learning with Depthwise Separable Convolutions, in: 2017 IEEE Conference on Computer Vision and Pattern Recognition (CVPR). IEEE, pp. 1800–1807. https://doi.org/10.1109/CVPR.2017.195

Collins, L., Hassani, H., Mokhtari, A., Shakkottai, S., 2022. FedAvg with Fine Tuning: Local Updates Lead to Representation Learning.

dos Anjos, C.E.M., Avila, M.R. V., Vasconcelos, A.G.P., Pereira Neta, A.M., Medeiros, L.C., Evsukoff, A.G., Surmas, R., Landau, L., 2021. Deep learning for lithological classification of carbonate rock micro-CT images. Comput. Geosci. 25, 971–983. https://doi.org/10.1007/s10596-021-10033-6

Fu, D., Su, C., Wang, W., Yuan, R., 2022. Deep learning based lithology classification of drill core images. PLoS One 17, e0270826. https://doi.org/10.1371/journal.pone.0270826

Galdames, F.J., Perez, C.A., Estévez, P.A., Adams, M., 2022. Rock lithological instance classification by hyperspectral images using dimensionality reduction and deep learning. Chemom. Intell. Lab. Syst. 224, 104538. https://doi.org/10.1016/j.chemolab.2022.104538

Glover, P.W.J., Mohammed-Sajed, O.K., Akyüz, C., Lorinczi, P., Collier, R., 2022. Clustering of facies in tight carbonates using machine learning. Mar. Pet. Geol. 144, 105828. https://doi.org/10.1016/j.marpetgeo.2022.105828

Goodfellow, I., Bengio, Y., Courville, A., 2017. Deep learning (adaptive computation and machine learning series). Cambridge Massachusetts 321–359.

He, K., Zhang, X., Ren, S., Sun, J., 2015. Deep Residual Learning for Image Recognition. https://doi.org/10.48550/arxiv.1512.03385

He, M., Gu, H., Xue, J., 2022. Log interpretation for lithofacies classification with a robust learning model using stacked generalization. J. Pet. Sci. Eng. 214, 110541. https://doi.org/10.1016/j.petrol.2022.110541

Huang, G., Liu, Z., Van Der Maaten, L., Weinberger, K.Q., 2017. Densely Connected Convolutional Networks, in: 2017 IEEE Conference on Computer Vision and Pattern Recognition (CVPR). IEEE, pp. 2261–2269. https://doi.org/10.1109/CVPR.2017.243

Hussain, M., Liu, S., Ashraf, U., Ali, M., Hussain, W., Ali, N., Anees, A., 2022. Application of Machine Learning for Lithofacies Prediction and Cluster Analysis Approach to Identify Rock Type. Energies 15, 4501. https://doi.org/10.3390/en15124501

Imamverdiyev, Y., Sukhostat, L., 2019. Lithological facies classification using deep convolutional neural network. J. Pet. Sci. Eng. 174, 216–228. https://doi.org/10.1016/j.petrol.2018.11.023

Izadi, H., Sadri, J., Bayati, M., 2017. An intelligent system for mineral identification in thin sections based on a cascade approach. Comput. Geosci. 99, 37–49. https://doi.org/10.1016/j.cageo.2016.10.010

Johari, A., Emami Niri, M., 2021. Rock physics analysis and modelling using well logs and seismic data for characterising a heterogeneous sandstone reservoir in SW of Iran. Explor. Geophys. 52, 446–461. https://doi.org/10.1080/08123985.2020.1836956

Liu, H., Wu, Y., Cao, Y., Lv, W., Han, H., Li, Z., Chang, J., 2020. Well Logging Based Lithology Identification Model Establishment Under Data Drift: A Transfer Learning Method. Sensors 20, 3643. https://doi.org/10.3390/s20133643

Liu, Z., Zhang, J., Li, Y., Zhang, G., Gu, Y., Chu, Z., 2022. Lithology Prediction of One-dimensional Residual Network Based on Regularization Constraints. J. Pet. Sci. Eng. 215, 110620. https://doi.org/10.1016/j.petrol.2022.110620

McCreery, E., Al-Mudhafar, W.J., 2017. Geostatistical Classification of Lithology Using Partitioning Algorithms on Well Log Data - A Case Study in Forest Hill Oil Field, East Texas Basin.

McMahan, H.B., Moore, E., Ramage, D., Hampson, S., Arcas, B.A. y, 2016. Communication-Efficient Learning of Deep Networks from Decentralized Data.

Nasirigerdeh, R., Bakhtiari, M., Torkzadehmahani, R., Bayat, A., List, M., Blumenthal, D.B., Baumbach, J., 2020. Federated Multi-Mini-Batch: An Efficient Training Approach to Federated Learning in Non-IID Environments.

Olivas, E.S., Guerrero, J.D.M., Sober, M.M., Benedito, J.R.M., Lopez, A.J.S., 2009. Handbook Of Research On Machine Learning Applications and Trends: Algorithms, Methods and Techniques - 2 Volumes. Information Science Reference - Imprint of: IGI Publishing, Hershey, PA.

Ren, Q., Zhang, H., Zhang, D., Zhao, X., Yan, L., Rui, J., 2022. A novel hybrid method of lithology identification based on k-means++ algorithm and fuzzy decision tree. J. Pet. Sci. Eng. 208, 109681. https://doi.org/10.1016/j.petrol.2021.109681

Sheller, M.J., Reina, G.A., Edwards, B., Martin, J., Bakas, S., 2018. Multi-Institutional Deep Learning Modeling Without Sharing Patient Data: A Feasibility Study on Brain Tumor Segmentation.

Simonyan, K., Zisserman, A., 2014. Very Deep Convolutional Networks for Large-Scale Image Recognition.

Szegedy, C., Ioffe, S., Vanhoucke, V., Alemi, A., 2016. Inception-v4, Inception-ResNet and the Impact of Residual Connections on Learning.

Taheri, A., Makarian, E., Manaman, N.S., Ju, H., Kim, T.-H., Geem, Z.W., RahimiZadeh, K., 2022. A Fully-Self-Adaptive Harmony Search GMDH-Type Neural Network Algorithm to Estimate Shear-Wave Velocity in Porous Media. Appl. Sci. 12, 6339. https://doi.org/10.3390/app12136339

Taheri, A., Ghashghaei, S., Beheshti, A. and RahimiZadeh, K., 2021. A novel hybrid DMHS-GMDH algorithm to predict COVID-19 pandemic time series. In *2021 11th international conference on computer engineering and knowledge (ICCKE)* (pp. 322-327). IEEE.

Tan, C.P., Ali, S.S., Yakup, M.H., Mustafa, M.A., Chidambaram, P., Mazeli, A.H., Hanifah, M.A., 2022. Mitigation of Geomechanical Risks for Long Term CO2 Geological Storage, in: Day 1 Mon, February 21, 2022. IPTC. https://doi.org/10.2523/IPTC-22451-MS

Wang, L., Guo, W., Chen, B., Yang, L., Bai, J., 2022. Lithology and fluid discrimination using combined seismic attributes with the constraint of rock physics: a case study from W field, South Sumatra Basin. Explor. Geophys. 1–15. https://doi.org/10.1080/08123985.2021.2021801

Xu, Z., Ma, W., Lin, P., Hua, Y., 2022. Deep learning of rock microscopic images for intelligent lithology identification: Neural network comparison and selection. J. Rock Mech. Geotech. Eng. 14, 1140–1152. https://doi.org/10.1016/j.jrmge.2022.05.009

Xu, Z., Shi, H., Lin, P., Liu, T., 2021. Integrated lithology identification based on images and elemental data from rocks. J. Pet. Sci. Eng. 205, 108853. https://doi.org/10.1016/j.petrol.2021.108853

Yamashita, R., Nishio, M., Do, R.K.G., Togashi, K., 2018. Convolutional neural networks: an overview and application in radiology. Insights Imaging 9, 611–629. https://doi.org/10.1007/s13244-018-0639-9

Yang, S., Xiao, W., Zhang, M., Guo, S., Zhao, J., Shen, F., 2022. Image Data Augmentation for Deep Learning: A Survey.

Zhang, P.Y., Sun, J.M., Jiang, Y.J., Gao, J.S., 2017. Deep Learning Method for Lithology Identification from Borehole Images. https://doi.org/10.3997/2214-4609.201700945